\documentclass{article}

\usepackage{arxiv}

\usepackage[utf8]{inputenc} 
\usepackage[T1]{fontenc}    
\usepackage{hyperref}       
\usepackage{url}            
\usepackage{booktabs}       
\usepackage{amsfonts}       
\usepackage{nicefrac}       
\usepackage{microtype}      
\usepackage{lipsum}
\usepackage{graphicx}

\usepackage{xcolor}
\usepackage{caption}
\hypersetup{
    colorlinks,
    linkcolor={blue!100!black},
    citecolor={blue!100!black},
}
\renewcommand{\normalsize}{\fontsize{10}{12.5}\selectfont}
\renewcommand{\small}{\fontsize{9}{12}\selectfont}
\title{Investigating myocardial infarction and its effects in patients with urgent medical problems using advanced data mining tools}

\author{
Tanya Aghazadeh \\
Department of Industrial Engineering\\
Azad University, Science and Research Branch \\
Tehran, Iran \\
\texttt{aghazadeh.tanya@gmail.com} \\
\And
Mostafa Bagheri \\
Department of Mechanical and Aerospace Engineering\\
University of California, San Diego\\
La Jolla, CA 92093 \\
\texttt{mstfbagheri@ucsd.edu} \\
}

\begin{document}
\maketitle
\begin{abstract}
In medical science, it is very important to gather multiple data on different diseases and one of the most important objectives of the data is to investigate the diseases. Myocardial infarction is a serious risk factor in mortality and in previous studies, the main emphasis has been on people with heart disease and measuring the likelihood of myocardial infarction in them through demographic features, echocardiography, and electrocardiogram. In contrast, the purpose of the present study is to utilize data analysis algorithms and compare their accuracy in patients with a heart attack in order to identify the heart muscle strength during myocardial infarction by taking into account emergency operations and consequently predict myocardial infarction. For this purpose, 105 medical records of myocardial infarction patients with fourteen features including age, the time of emergency operation, Creatine Phosphokinase (CPK) test, heart rate, blood sugar, and vein are gathered and investigated through classification techniques of data analysis including random decision forests, decision tree, support vector machine (SVM), k-nearest neighbor, and ordinal logistic regression. Finally, the model of random decision forests with an accuracy of 76\% is selected as the best model in terms of the mean evaluation indicator. Also, seven features of the creatine Phosphokinase test, urea, white and red blood cell count, blood sugar, time, and hemoglobin are identified as the most effective features of the ejection fraction variable.
\end{abstract}


\section{Introduction}
Knowledge discovery in a database is a process that involves several distinct steps. Data analysis is the core of this process, which leads to the discovery of latent but useful knowledge from vast databases \cite{srinivas2010applications}. However, these datasets are valueless themselves and should be analyzed and turned into information and even knowledge to be valuable \cite{subbalakshmi2011decision}. Data analysis is the process of finding unknown models and trends and utilizing information to build predictive models \cite{bhatla2012analysis}, which has the potential to create a knowledge-rich environment and can significantly contribute to the quality of clinical decisions \cite{das2009effective}. The application of data analysis has been proven in various medical fields such as disease diagnosis \cite{lavravc1999selected}, finding hidden pattern \cite{begdache2019assessment}, and investigating new trends.

The medical and health field is important sector of industrial societies. The volume of medical data is increasing day by day and physicians can obtain valuable information about diseases and their relationship with other disease-causing factors \cite{soni2011predictive}. Data analysis is also used in the field of myocardial infarction as a tool to convert raw data into knowledge and information. A heart attack (myocardial infarction) occurs when the blood flow stops at one part of the heart and often occurs due to the creation of clots when coronary arteries that pump blood to the heart are gradually blocked by a network of cells, fat, and cholesterol called plaque. The blood flow passing through these blocked arteries can form a clot, and if this clot completely cuts off blood flow, the part of the heart muscle that feeds the artery will be damaged and eventually die \cite{thygesen2007universal}.
 
In recent years, utilizing new approaches in biomedical researches has become very important by the worldwide prevalence of cardiovascular diseases. In this regard, data analysis is also a helpful tool for acquiring such knowledge. The healthcare environment is rich in information but poor in knowledge \cite{boo2012cardiovascular}. The cooperation of computer and medical experts can provide a new solution for analyzing medical data and obtaining useful and applied models that are medical data analysis. Therefore, it is significantly important for developing predictive models to utilize data analysis techniques in order to identify high-risk individuals and decrease the side-effects of disease \cite{soni2011predictive}.

Using data analysis process to extract knowledge from a large volume of data related to disease histories and medical records of individuals can lead to the identification of the rules governing the creation, growth, and acceleration of diseases and by considering environmental factors, provide health professionals and practitioners with valuable information to identify the causes of disease occurrence, diagnosis, prediction, and treatment. The result is an increase in community well-being and peace. Therefore, the complexity of medical information and the existence of data analysis tools on medical and health data are considered important. It would be possible to take necessary measures in order to prevent a person from having a heart attack and other disasters in the case of having certain variables about him/her \cite{shouman2012using}. The main objective of the present study was to identify the important effective factors of heart muscle strength in relation to heart failure as well as myocardial infarction in Iran.

\section{Research Objectives}
\label{sec:headings}
The objective of the present study is to predict Ejection Fraction (EF) through the evaluation of medical records and Hospital Information System (HIS) data in addition to information about Tehran emergency operations in relation to patients with myocardial infarction transferred from one of Tehran's Emergency Medical Services (EMS) to Tehran Public Hospitals in the first half of 2018. We find the effective features and investigate their effect on EF and consequently, provide the condition of helping patients with myocardial infarction in the shortest possible time.

\section{Methodology}
CRISP-DM process is used in myocardial infarction predictors as a standard data analysis methodology to make data analysis models. This methodology provides a process model for data analysis that is an overview of the life cycle of any data analysis project and includes the steps corresponding to a project, its related tasks, and the relationship between these tasks \cite{chapman2000crisp}. The process consists of six steps including business understanding, data understanding, data preparation, modeling, evaluation, and development. The sequence of these steps is not straightforward and moving back and forth between different steps is always required. The output of each step specifies what step or the task of a particular step should be performed as the next one.

\subsection{Data Source}
The data of the present study was gathered from four thousand myocardial infarction patients transferred to the public hospitals of Tehran by branch 247 of EMS as well as 351 patients related to data of HIS System. The gathered data from these patients included emergency data about the time of operation, the study of medical records of patients, and data available in the Tehran HIS system as a database during the first half of 2018. At first, the names of patients in the list of EMS were matched with the names of patients in the list of HIS, and 105 patients with myocardial infarction shared in both lists were obtained.

Medical records of 105 final patients were investigated individually and patient features were recorded. The features were categorized into four main groups: 1) Patient demographic features, 2) Patient laboratory features, 3) Patient clinical examination features, and 4) Time and type of emergency operations. The number of variables was obtained was 64 features with completing the process of making aggregated data table.

Firstly, Imputation was implemented using packages available in the R software to manage lost or missing data. Then, the data was balanced through an upsampling technique to avoid bias and this function generated some random observations in order to model the data with more observations (i.e. 126 observations). Finally, the step of important feature selection methods and modeling was carried out on data.

Feature selection methods have turned into one of the most important elements of the learning process in order to face high volume data. The feature selection algorithm of the present study is a random decision forest algorithm. This algorithm is an observer-based learning technique in which many random decision trees are generated during training. It is resistant to noise and error and capable of dealing with unbalanced and even lost data. For this purpose, a feature selection method of the present study is in the form of Recursive Feature Elimination (RFE\footnote{Recursive Feature Elimination (RFE) is a feature selection method that fits a model and removes the weakest feature (or features) until the specified number of features is reached.}). 

In feature selection using the result of the output of R code, the important features were identified by drawing and interpreting the graph of mean squared error or Root-mean-square deviation (RMSE) so that the RMSE indicated the difference between the value predicted by the statistical estimator and the real value. RMSE is used to evaluate the model and is a good tool for comparing prediction errors by a dataset (a lower level of error means higher utility). Finally, fourteen important and effective final features of predicting myocardial infarction were entered into the model in the first step, and have been presented in Figure \ref{fig:list_1}.

The second step was related to the prediction of the most important operational features affecting the EF variable without taking into account other variables such as demographic, laboratory, and clinical features. In this step, six final features were entered into the model. These six final features were selected among 24 initial features of the feature selection step by the mentioned method, and have been presented in Figure \ref{fig:list_2}.

\begin{figure}
    \small
    \fbox{
        \begin{minipage}{0.9\linewidth}
        \textbf{Predictable attribute}
        \vspace{0.5cm}
        
        Ejection Fraction
        \begin{itemize}
            \item value0: 50-70\%: Normal and no heart attack
            \item value1: 36-49\%: Below Normal and moderate
            \item value2: <35\%: Low (severe heart attack)
        \end{itemize}
        \vspace{0.5cm}
        
        \textbf{Input attributes}
        \begin{enumerate}
            \item Age (patient age in year)
            \item LAD (left anterior descending; value1: Yes, value0: No)
            \item W.B.C (White Blood Cells)
            \item R.B.C (Red Blood Cells)
            \item B.U.N (Blood Urea Nitrogen)
            \item HB (Hemoglobin)
            \item CPK (Creatine Phosphokinase)
            \item CPK-MB(Creatine Phosphokinase/MB)
            \item PR (Heart Rate)
            \item BS (Blood Sugar)
            \item Time X1+X2 (continuous)
            \item Time X1+X2+X3+X4 (continuous)
            \item Time X2+X3 (continuous)
            \item Heart norm sound (value1: Yes; value0: No)
        \end{enumerate}
        
        X1 = The time difference between an emergency call and pain onset
        
        X2 = The time difference between arrival at the patient's place and an emergency contact
        
        X3 = The time difference between arrival at the cath lab and hospital admission
        
        X4 = The time difference between hospital admission and arrival at the patient's place
        
        \end{minipage}
    }
    \caption{List of attributes used for heart attack prediction in step 1}
    \label{fig:list_1}
\end{figure}

\begin{figure}[!h]
    \small
    \fbox{
        \begin{minipage}{0.9\linewidth}
        \textbf{Predictable attribute}
        \vspace{0.5cm}
        
        1.Ejection fraction
        \begin{itemize}
            \item value0: 50-70\%: Normal and no heart attack
            \item value1: 36-49\%: Below Normal and moderate
            \item value2: <35\%: Low (severe heart attack)
        \end{itemize}
        \vspace{0.5cm}
        
        \textbf{Input attributes}
        \begin{enumerate}
            \item Time X1+X2 (continuous)
            \item Time X1+X2+X3+X4 (continuous)
            \item Time X2+X3 (continuous)
            \item Time X1+X2+X3 (continuous)
            \item Heart norm sound (value1: Yes; value0: No)
            \item Time FMC - Onset (continuous)

        \end{enumerate}
        
        X1 = The time difference between an emergency call and pain onset
        
        X2 = The time difference between arrival at the patient's place and an emergency contact
        
        X3 = The time difference between arrival at the cath lab and hospital admission
        
        X4 = The time difference between hospital admission and arrival at the patient's place
        
        Fmc.onset = FMC - ONSET = The time difference between an emergency call and a pain onset

        \end{minipage}
    }
    \caption{List of operational attributes used for heart attack prediction in step 2}
    \label{fig:list_2}
\end{figure}

\subsection{Mining Models}
In the present study, observer-based machine learning algorithms including random decision forests, k-nearest neighbors, decision tree, ordinal logistic regression, and support vector machine were used to classify the data. The models were developed by the RStudio programming language, which is a reliable and accurate tool in academic research.

\subsection{Confusion Matrix}
A confusion matrix is a table to illustrate the algorithm's performance and accuracy in classification, which has been shown in Tables \ref{tab:1-1} to \ref{tab:1-5} for five classification methods with fourteen attributes. In our study, we have three classes, therefore we have a 3 by 3 confusion matrix.

\begin{table}[!h]
    \small
    \caption{Confusion matrix for Random Decision Forest}
    \begin{center}
        \begin{tabular}{ | c | c | c | c | } 
            \hline
            & Normal & Low heart failure & High heart failure \\ 
            \hline
            Normal & 28 & 10 & 4 \\ 
            \hline
            Low heart failure & 5 & 26 & 11 \\ 
            \hline
            High heart failure & 8 & 7 & 27 \\ 
            \hline
        \end{tabular}
    \end{center}
    \label{tab:1-1}
\end{table}

\begin{table}[!h]
    \small
    \caption{Confusion matrix for Support Vector Machine}
    \begin{center}
        \begin{tabular}{ | c | c | c | c | } 
            \hline
            & Normal & Low heart failure & High heart failure \\ 
            \hline
            Normal & 10 & 29 & 3 \\ 
            \hline
            Low heart failure & 3 & 37 & 2 \\ 
            \hline
            High heart failure & 2 & 20 & 20 \\ 
            \hline
        \end{tabular}
    \end{center}
    \label{tab:1-2}
\end{table}

\begin{table}[!h]
    \small
    \caption{Confusion matrix for Decision Tree}
    \begin{center}
        \begin{tabular}{| c | c | c | c | } 
            \hline
            & Normal & Low heart failure & High heart failure \\ 
            \hline
            Normal & 23 & 8 & 11 \\ 
            \hline
            Low heart failure & 8 & 22 & 12 \\ 
            \hline
            High heart failure & 3 & 11 & 28 \\ 
            \hline
        \end{tabular}
    \end{center}
    \label{tab:1-3}
\end{table}

\begin{table}[!h]
    \small
    \caption{Confusion matrix for K-NN}
    \begin{center}
        \begin{tabular}{ | c | c | c | c | } 
            \hline
            & Normal & Low heart failure & High heart failure \\ 
            \hline
            Normal & 20 & 18 & 4 \\ 
            \hline
            Low heart failure & 6 & 34 & 2 \\ 
            \hline
            High heart failure & 5 & 13 & 24 \\ 
            \hline
        \end{tabular}
    \end{center}
    \label{tab:1-4}
\end{table}

\begin{table}[!h]
    \small
    \caption{Confusion matrix for Ordinal Logistic Regression}
    \begin{center}
        \begin{tabular}{ | c | c | c | c | } 
            \hline
            & Normal & Low heart failure & High heart failure \\ 
            \hline
            Normal & 23 & 8 & 3 \\ 
            \hline
            Low heart failure & 8 & 22 & 11 \\ 
            \hline
            High heart failure & 11 & 12 & 28 \\ 
            \hline
        \end{tabular}
    \end{center}
    \label{tab:1-5}
\end{table}

\section{Evaluating the efficiency of model}
\begin{figure}
    \centering
    \includegraphics[width=0.62\linewidth]{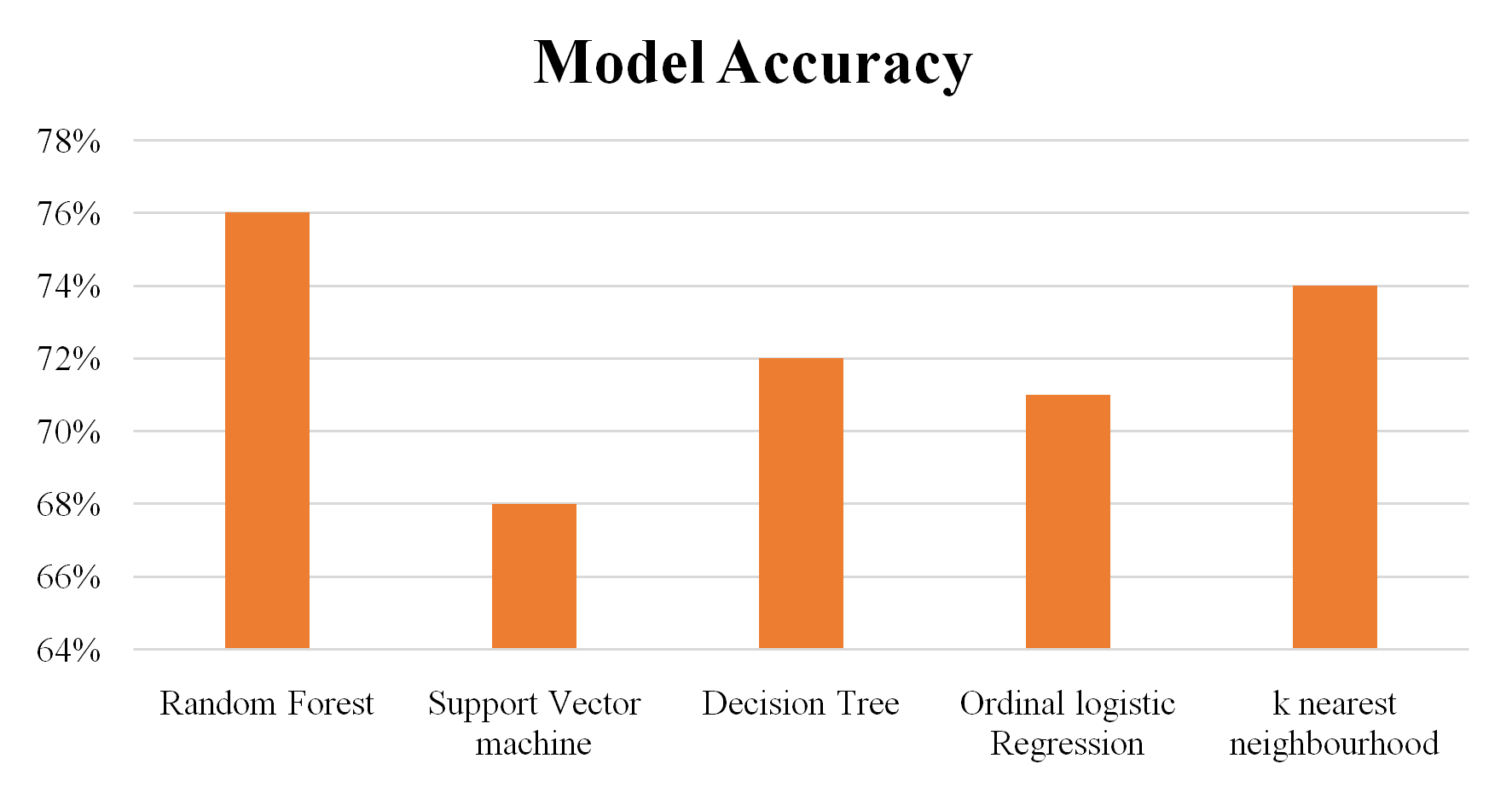}
    \caption{Model accuracy results}
    \label{fig:acc}
\end{figure}

\setcounter{footnote}{0}
The accuracy of a classification algorithm is the most important criterion for determining its efficiency. This criterion calculates the overall accuracy of a classifier and indicates what percentage of the whole set of test records has been correctly classified by the classifier. Model’s average accuracy is considered in final decision making. A test set independent from the training set is considered to measure the performance and accuracy of models. Each algorithm is studied through two sets of learning and test and the target variable for the understudy dataset are estimated and compared with the real value. The five understudy models in the ten-fold cross-validation\footnote{When a specific value for k is chosen, it may be used in place of k in the reference to the model, such as k=10 becoming ten-fold cross-validation. Cross-validation is primarily used in applied machine learning to estimate the skill of a machine learning model on unseen data} method in relation to hospital data are random decision forest, k-nearest neighbors, decision tree, ordinal logistic regression, and support vector machine were ranked in priority with an average accuracy of 76\%, 74\%, 72\%, 71\%, and 68\%, respectively; the results have been presented in Figure \ref{fig:acc}.

Among the models, the model of random decision forest had the highest average accuracy (76\%) and highest efficiency (65\%) and was selected as the final model, and the models of k-nearest neighbors, support vector machine, decision tree, and Ordinal logistic regression were ranked in the next places with 64\%, 63\%, 58\%, and 57\%, respectively. The detailed performance measures (precision, recall, F-score\footnote{F-score indicator is a combination of two criteria of model accuracy and model recall to evaluate the classifier algorithms, in other words, a harmonic mean of accuracy and recall.}, and G-score\footnote{G-score indicator is an arithmetic mean of accuracy and recall.}) for five classification models have been presented in Tables \ref{tab:2-1} to \ref{tab:2-5}. Note that the F-score indicator and the G-score are calculated for all five models. Among five understudy models, the model of random decision forests had the highest values of G-score (63\%) and F-score (65\%).

\begin{table}[!h]
    \small
    \caption{Detailed performance measures for Random Decision Forest}
    \begin{center}
        \begin{tabular}{ | c | c | c | c | c |} 
            \hline
            & Precision & Recall & F-score & G-score \\ 
            \hline
            Normal & 68\% & 66\% & 67\% & 67\% \\ 
            \hline
            Low heart failure & 60\% & 61\% & 61\% & 61\% \\ 
            \hline
            High heart failure & 64\% & 64\% & 64\% & 64\% \\ 
            \hline
            Metrics (Average) & 65\% & 64\% & 65\% & 63\% \\
            \hline
            Accuracy(Average) & \multicolumn{4}{|c|}{76\%} \\
            \hline
        \end{tabular}
    \end{center}
    \label{tab:2-1}
\end{table}

\begin{table}[!h]
    \small
    \caption{Table detailed performance measures for SVM}
    \begin{center}
        \begin{tabular}{ | c | c | c | c | c |} 
            \hline
            & Precision & Recall & F-score & G-score \\ 
            \hline
            Normal & 66\% & 23\% & 35\% & 39\% \\ 
            \hline
            Low heart failure & 43\% & 88\% & 57\% & 61\% \\ 
            \hline
            High heart failure & 80\% & 47\% & 60\% & 62\% \\ 
            \hline
            Metrics (Average) & 63\% & 53\% & 50\% & 54\% \\
            \hline
            Accuracy(Average) & \multicolumn{4}{|c|}{68\%} \\
            \hline
        \end{tabular}
    \end{center}
    \label{tab:2-2}
\end{table}

\begin{table}[!h]
    \small
    \caption{Detailed performance measures for Decision Tree}
    \begin{center}
        \begin{tabular}{ | c | c | c | c | c |} 
            \hline
            & Precision & Recall & F-score & G-score \\ 
            \hline
            Normal & 67\% & 54\% & 60\% & 61\% \\ 
            \hline
            Low heart failure & 53\% & 52\% & 53\% & 53\% \\ 
            \hline
            High heart failure & 54\% & 66\% & 60\% & 60\% \\ 
            \hline
            Metrics (Average) & 58\% & 57\% & 58\% & 57\% \\
            \hline
            Accuracy(Average) & \multicolumn{4}{|c|}{72\%} \\
            \hline
        \end{tabular}
    \end{center}
    \label{tab:2-3}
\end{table}

\begin{table}[!h]
    \small
    \caption{Detailed performance measures for Ordinal Logistic Regression}
    \begin{center}
        \begin{tabular}{ | c | c | c | c | c |} 
            \hline
            & Precision & Recall & F-score & G-score \\ 
            \hline
            Normal & 54\% & 67\% & 60\% & 61\% \\ 
            \hline
            Low heart failure & 52\% & 53\% & 53\% & 53\% \\ 
            \hline
            High heart failure & 66\% & 54\% & 60\% & 60\% \\ 
            \hline
            Metrics (Average) & 57\% & 58\% & 59\% & 56\% \\
            \hline
            Accuracy(Average) & \multicolumn{4}{|c|}{71\%} \\
            \hline
        \end{tabular}
    \end{center}
    \label{tab:2-4}
\end{table}

\begin{table}[!h]
    \small
    \caption{Detailed performance measures for K-NN}
    \begin{center}
        \begin{tabular}{ | c | c | c | c | c |} 
            \hline
            & Precision & Recall & F-score & G-score \\ 
            \hline
            Normal & 64\% & 47\% & 54\% & 55\% \\ 
            \hline
            Low heart failure & 52\% & 80\% & 63\% & 65\% \\ 
            \hline
            High heart failure & 80\% & 58\% & 66\% & 67\% \\ 
            \hline
            Metrics (Average) & 64\% & 62\% & 61\% & 63\% \\
            \hline
            Accuracy(Average) & \multicolumn{4}{|c|}{74\%} \\
            \hline
        \end{tabular}
    \end{center}
    \label{tab:2-5}
\end{table}

In the second step of modeling, six final features were modeled only through operational variables including emergency operation time. Table \ref{tab:result} represents the results obtained from prioritization of the models along with their evaluation metrics, i.e. model’s average accuracy. Again, the model of random decision forests was superior to other models with an average accuracy of 70\%. The confusion matrix of final model has been represented in Table \ref{tab:best_model_m}. Other indicators of evaluating accuracy and efficiency of the final model have been represented in Table \ref{tab:best_model}.

\begin{table}[!h]
    \small
    \caption{Results obtained from prioritization of the models accuracy}
    \begin{center}
        \begin{tabular}{ | c | c |} 
            \hline
            Model Name & Mean Accuracy \\ 
            \hline
            Random Decision Forest & 70\% \\ 
            \hline
            K-nearest neighbors & 65\% \\ 
            \hline
            Ordinal logistic regression & 64\% \\ 
            \hline
            Decision Tree & 63\% \\
            \hline
            Support Vector Machine & 60\% \\
            \hline
        \end{tabular}
    \end{center}
    \label{tab:result}
\end{table}

\begin{table}[!h]
    \small
    \caption{Confusion matrix obtained for the best model (random decision forest) through operational variables}
    \begin{center}
        \begin{tabular}{ | c | c | c | c |} 
            \hline
            & Normal & Low heart failure & High heart failure \\ 
            \hline
            Normal & 33 & 5 & 4 \\ 
            \hline
            Low heart failure & 10 & 18 & 14 \\ 
            \hline
            High heart failure & 8 & 13 & 21 \\ 
            \hline
        \end{tabular}
    \end{center}
    \label{tab:best_model_m}
\end{table}

\begin{table}[!h]
    \small
    \caption{Detailed performance measures for the best model (random decision forest)}
    \begin{center}
        \begin{tabular}{ | c | c | c | c | c |} 
            \hline
            & Precision & Recall & F-score & G-score \\ 
            \hline
            Normal & 64\% & 78\% & 70\% & 71\% \\ 
            \hline
            Low heart failure & 50\% & 42\% & 46\% & 46\% \\ 
            \hline
            High heart failure & 53\% & 50\% & 51\% & 51\% \\ 
            \hline
            Metrics (Average) & 56\% & 57\% & 56\% & 56\% \\
            \hline
            Accuracy(Average) & \multicolumn{4}{|c|}{70\%} \\
            \hline
        \end{tabular}
    \end{center}
    \label{tab:best_model}
\end{table}

\section{Results}
\begin{figure}[h]
    \centering
    \includegraphics[width=0.65\linewidth]{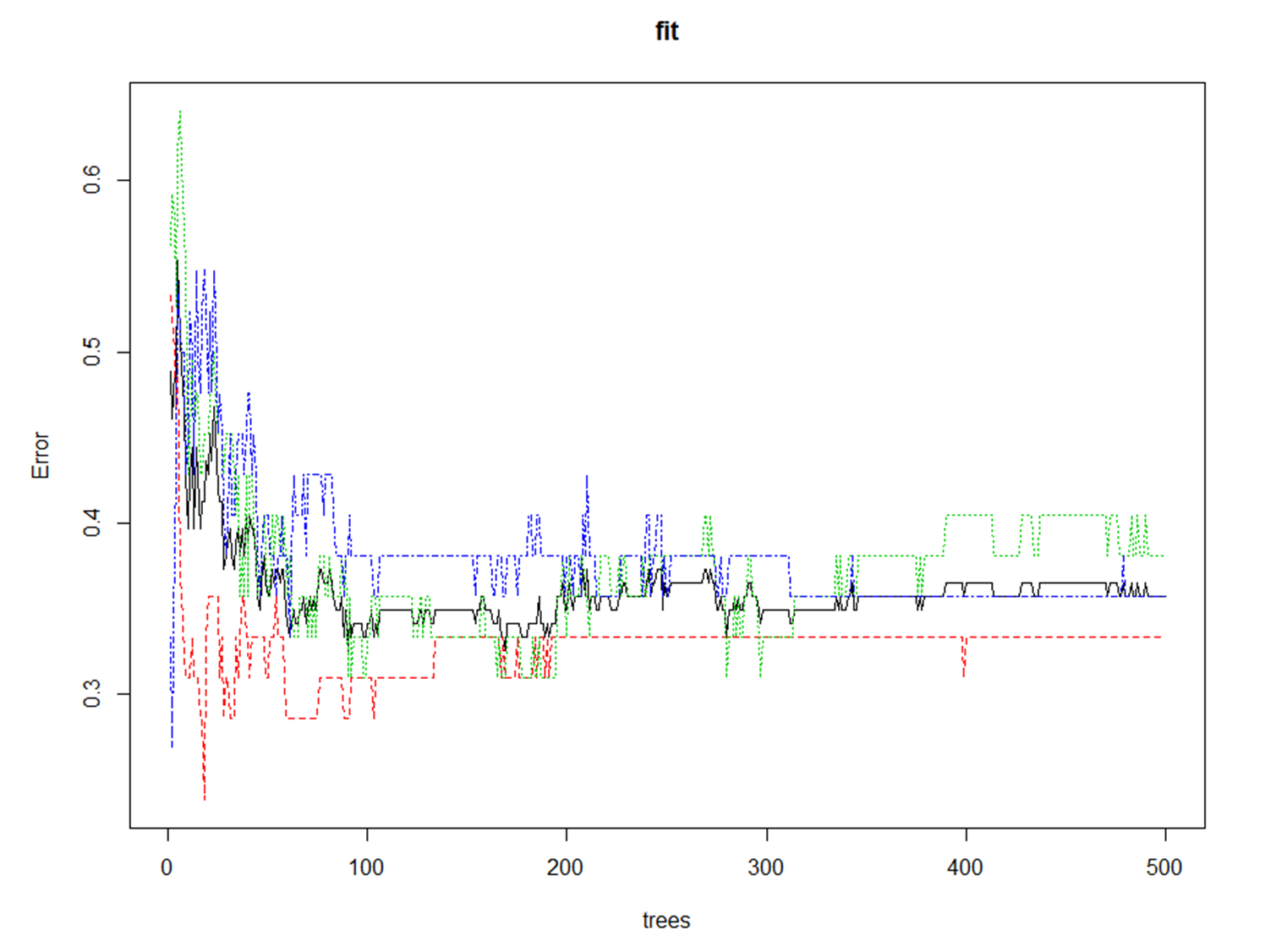}
    \caption{Random Decision Forest error plot}
    \label{fig:error}
\end{figure}

As mentioned before, the model of random decision forests was selected as the final model, which its error plot has been presented in Figure \ref{fig:error}. As can be seen, the error rate was initially high with an increase in the number of trees up to 500 and then gradually decreased and became almost constant at the end.

\begin{figure}[!h]
    \centering
    \includegraphics[width=0.5\linewidth]{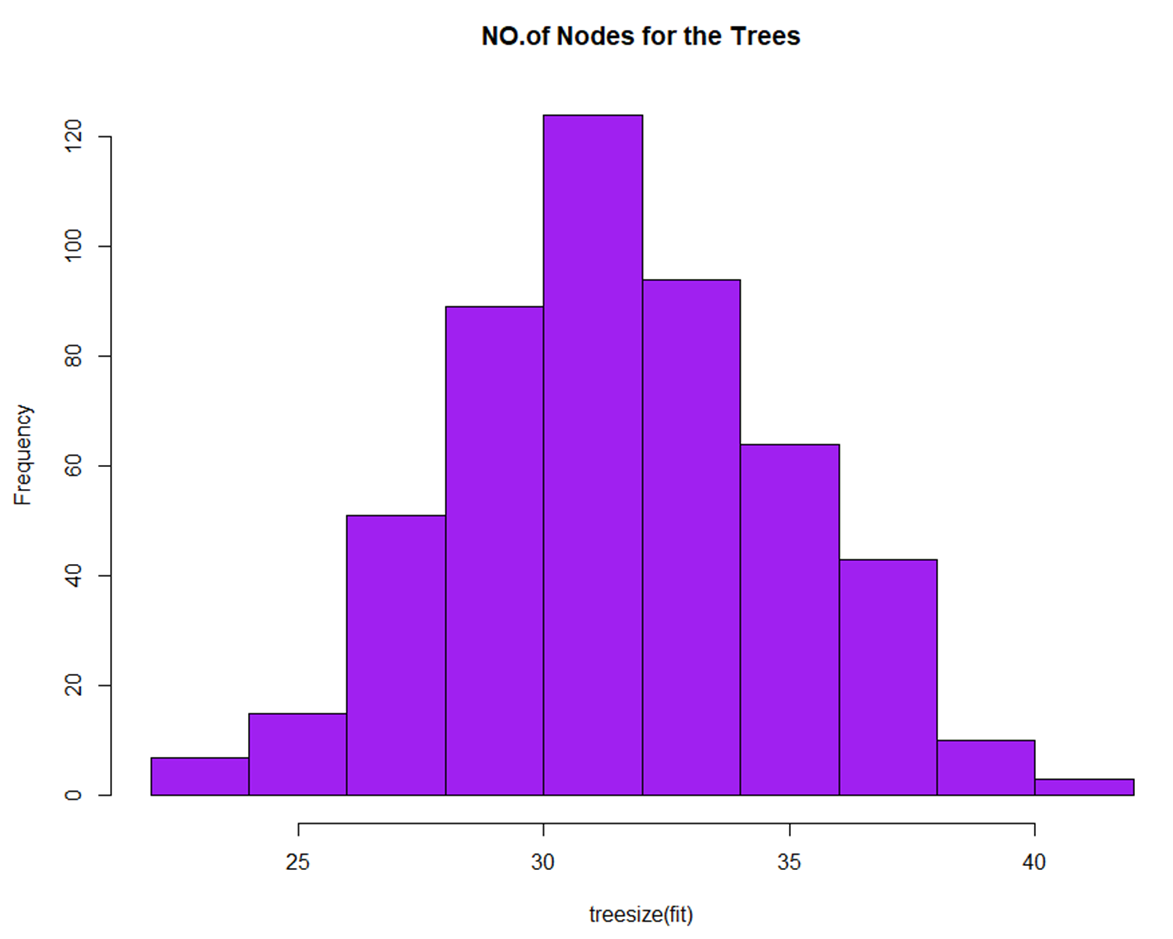}
    \caption{Random Decision Forest histogram plot}
    \label{fig:RF_h}
\end{figure}

The histogram diagram of the number of decision tree nodes in the random decision forest model has been presented in Figure \ref{fig:RF_h}. This graph represents the distribution nodes in each of the 500 available trees. As can be seen, the longest rod is near the number 120, indicating that there are approximately 32 nodes in the 120 available trees. On the other hand, a few trees at the beginning of the graph (approximately 20 trees) have less than 20 nodes and fewer trees at the bottom of the graph have more than 40 nodes. Finally, the distribution of nodes in 500 trees is between 25 and more than 35 nodes, however, most trees have 32 nodes.

\begin{figure}[!h]
    \centering
    \includegraphics[width=0.65\linewidth]{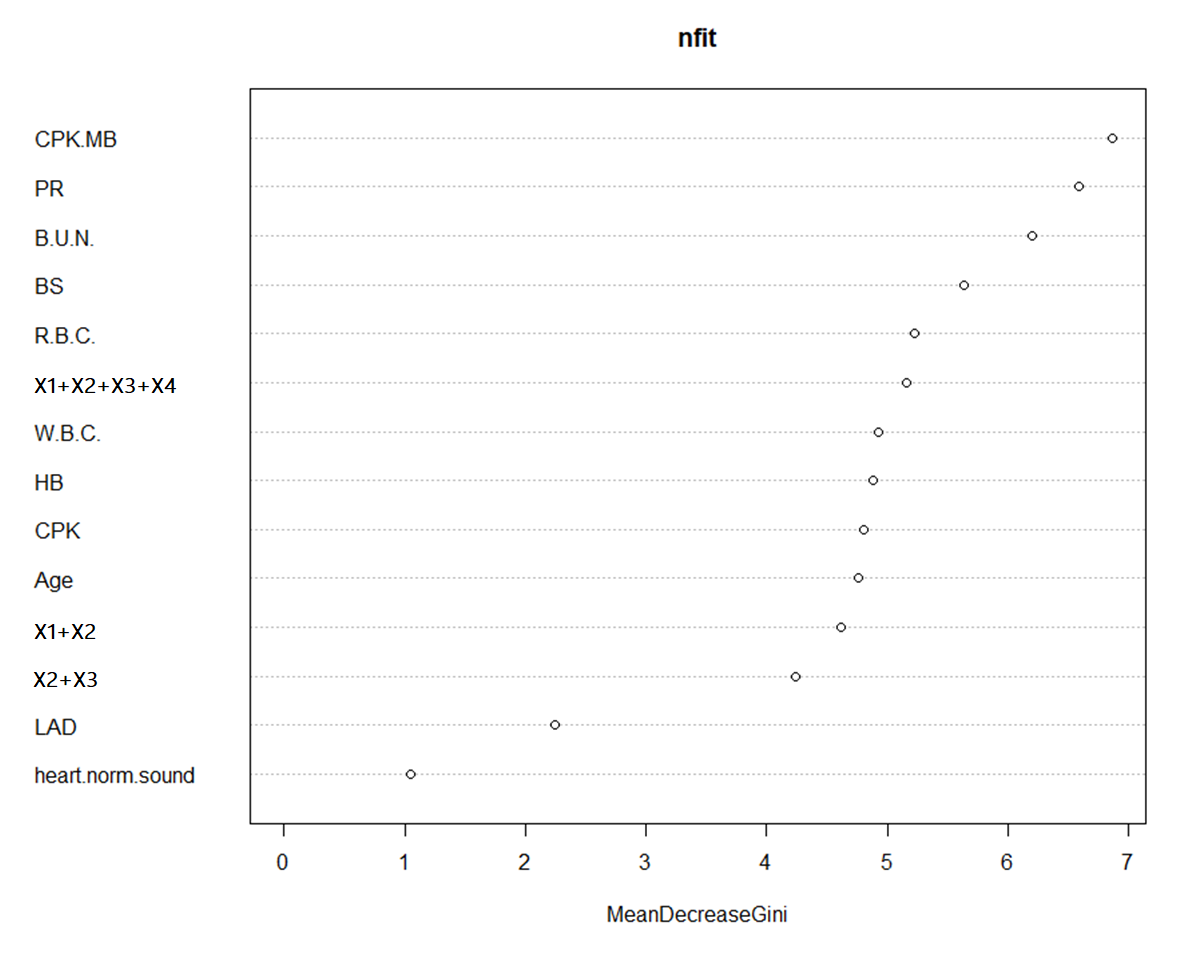}
    \caption{The most important effective features for predicting EF step 1}
    \label{fig:gini}
\end{figure}

In continuing the most important part of the study has been described i.e. identifying the most important effective features for predicting EF variable after modeling in R software and selecting the random decision forests as the final model. The most important obtained features have been presented in the diagram of Figure \ref{fig:gini}.

As can be seen in Figure \ref{fig:gini}, the most important features of the model in the order of importance included CPK-MB creatine Phosphokinase test, heart beat (PR), blood urea nitrogen (B.U.N ), blood sugar (B.S), red blood cell (R.B.C) count, sum of time difference from the moment of pain onset to arrival of cath lab X1+X2+X3+X4 and W.B.C white blood cell count, which were used to the prediction of EF variable. In continuing the effectiveness of each of them was investigated in accordance with the valid medical references and consultation with a cardiologist.

\subsection{The effect of final features on the ejection fraction variable}
In this section,7 final features affecting EF in myocardial infarction patients have been investigated.CPK-MB creatine Phosphokinase test was the first important effective feature of ejection fraction variable and according to the study of Brian Gibler et al. \cite{gibler1990early}, this test can make the diagnosis faster with a higher sensitivity to the heart attack. Also according to the study conducted by Javier Copi et al. \cite{copie1996predictive} with a two-year follow-up of patients, the patient’s pulsation rate (PR) and respiratory rate (RR) per minute have an effect on EF or heart function of patient and patient's heart function in pumping blood is declined with increase in heart pulsation rate and decrease in respiratory rate. The test of B.U.N or blood urea nitrogen was also identified as an effective feature of a heart attack.

The test indicates that increased B.U.N can cause kidney failure and heart attack. In 2005, a group of researchers \cite{trialists2005effects}, investigated the effect of blood pressure (BP) and blood sugar (BS) on the death of cardiovascular patients in diabetic and non-diabetic conditions with changes in blood pressure. The obtained results showed that the probability of a heart attack is increased with an increase in blood pressure (BP) and blood sugar (BS). About the feature of time as one of the main concerns of the present study, it can be said that decreasing the time difference between patient's onset of pain and patient's arrival to cath lab of the hospital has a significant effect on saving the lives of heart attack patients. We think that this would be possible through training of technicians, training of public by the media and traffic control. Also, high white blood cell count W.B.C, as well as high red blood cell count R.B.C, can be a sign of heart attack occurred in the patient.

\begin{figure}[!ht]
    \centering
    \includegraphics[width=0.65\linewidth]{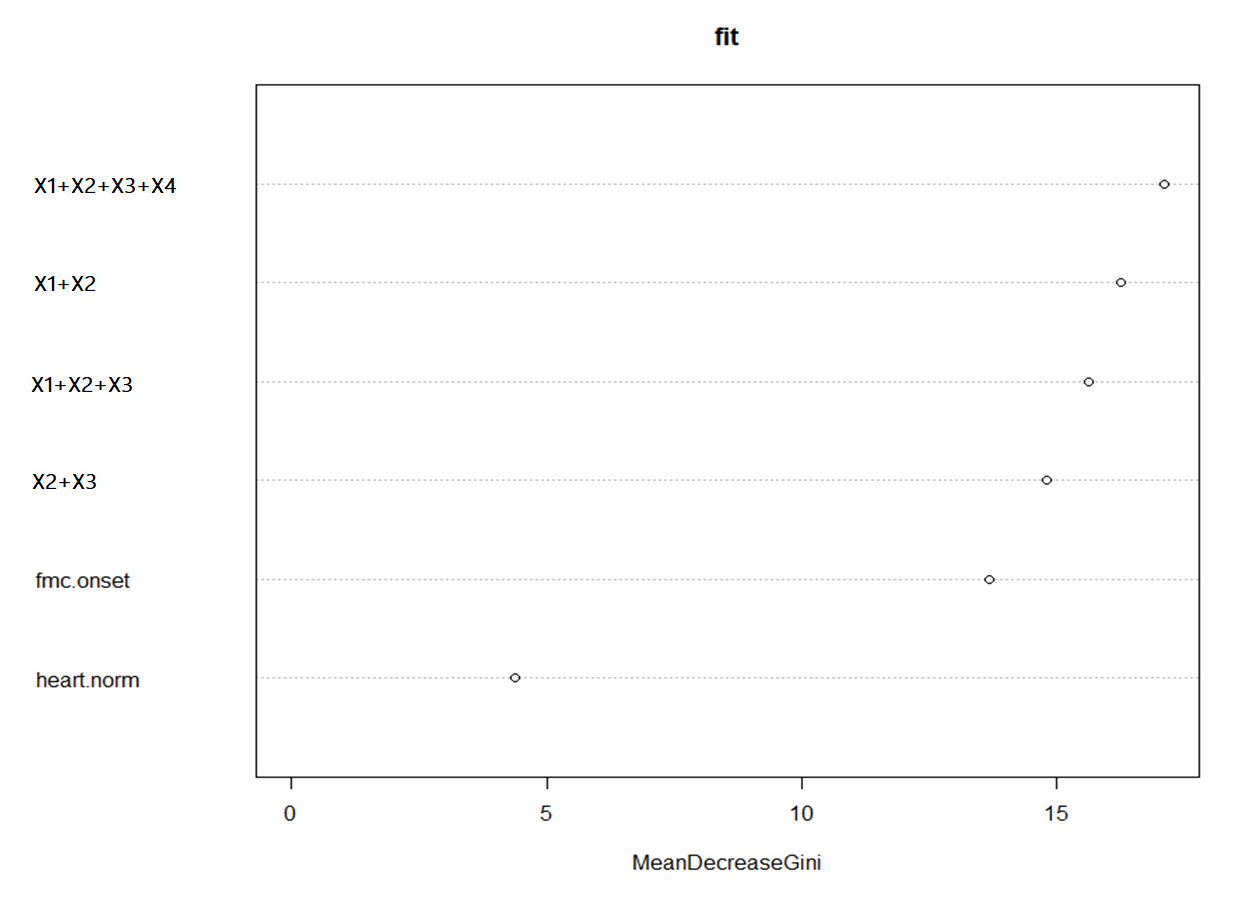}
    \caption{The most important effective features for predicting EF step 2}
    \label{fig:gini_2}
\end{figure}

As can be seen in Figure \ref{fig:gini_2}, the final output of the present study in the second step included five important variables in relation to ejection fraction as follow:
\begin{itemize}
    \item The total of the time difference between an emergency call and pain onset (X1), the time difference between arrival at the emergency place and emergency contact (X2), the time difference between arrival at the cath lab and hospital admission (X3), and the time difference between hospital admission and arrival at the medical emergency place (X4)
    \item The total of the time difference between an emergency call and pain onset (X1) and the time difference between arrival at the emergency place and emergency contact (X2)
    \item The total of the time difference between an emergency call and pain onset (X1), the time difference between arrival at the emergency place and emergency contact (X2), and the time difference between arrival at the cath lab and hospital admission (X3)
    \item The total of the time difference between arrival at the emergency place and emergency contact (X2) and time difference between arrival at the cath lab and hospital admission (X3) 
    \item The time difference between an emergency call and pain onset
\end{itemize}
and the first variable is the most important one.

\section{Conclusion}
Cardiovascular diseases have been considered a major global challenge. The treatment of this disease imposes a high cost on the health system of countries. The investigations indicate a high prevalence of cardiovascular disease and its contributing factors in Iran and the development of a comprehensive program should be considered for the prevention of these diseases as soon as possible. Myocardial infarction (MI) or acute myocardial infarction (AMI) is the interruption of blood flow and failure of blood flow to reach the heart due to a slow or abrupt reduction of coronary artery blood flow.

Ischemic heart diseases are divided into two groups: heart infection and heart failure. In myocardial infarction or heart attack, permanent and irreversible cell death and death in part of the heart muscle (myocardium) occur due to interruption of blood flow and the occurrence of severe ischemia. Heart failure means a severe decline in heart function and the power of contraction, which results in insufficient pumping of blood. Heart failure occurs when the heart is unable to effectively pump blood to all parts of the body resulting in insufficient oxygen supply, and subsequently enlarging of heart and thickness of muscle fibers and heart rate to compensate.

The EF or heart strength above 55 and 60 is considered a normal heart condition, and a lower EF (about 30 or less) indicates a heart problem. Any motion abnormalities in the wall or valves of the heart can indicate the occurrence of a cardiovascular event. According to the obtained results, all five understudy data analysis models had a proper performance in analyzing the strength of heart muscles in myocardial infarction and among them, the random decision forest algorithm showed the best performance.

According to 105 hospital records and HIS system data and information on emergency operations of acute myocardial infarction patients who were transferred to one of the public hospitals of Tehran by EMS during the first half of 2018 as well as by utilizing data analysis techniques and R software for analyzing data and also by carefully analyzing the features and nature of the data including sequential and classes of ejection fraction, five data analysis models including random decision forest, support vector machine, decision tree, sequential logistic regression, and k-nearest neighbor were identified to predict the function of heart muscles in relation to the level of heart failure caused by myocardial infarction and the effect of other important variables on ejection fraction of “left ventricular pumping rate'' was also investigated.

After the formation of confusion matrix through ten-fold cross-validation method and its analysis and comprising the criteria of precision, accuracy, harmonic and geometric mean, the criterion of average accuracy was finally selected for the model as a reliable criterion and the model of random decision forests with an average accuracy of 76\% in ten-fold cross-validation method has a better performance compared to other models implemented on the data and was selected as the best model.

\bibliographystyle{IEEEtran}
\bibliography{references}

\end{document}